%% file: main.tex
\pdfoutput=1
\documentclass[10pt,twocolumn,letterpaper]{article}
\usepackage{authblk}
\usepackage{cvpr}
\usepackage{times}
\usepackage{epsfig}
\usepackage{graphicx}
\usepackage{amsmath}
\usepackage{amssymb}
\usepackage{microtype}
\usepackage[flushleft, alwaysadjust]{paralist}
\makeatletter
\providecommand\@enum@widestlabel{7}
\makeatother

\usepackage[pagebackref=true,breaklinks=true,letterpaper=true,colorlinks,bookmarks=false]{hyperref}

\input{macros.tex}

\cvprfinalcopy %

\ifcvprfinal\fi
\begin{document}

\title{\ourpapertitle}

\author{Orly Liba}
\author{Longqi Cai}
\author{Yun-Ta Tsai}
\author{Elad Eban}
\author{Yair Movshovitz-Attias}
\author{Yael Pritch}
\author{Huizhong Chen}
\author{Jonathan T. Barron}
\affil{Google Research}

\affiliation{
\institution{Google Research}
\streetaddress{1600 Amphitheatre Parkway}
\city{Mountain View}
\state{CA}
\postcode{94043}
}

\maketitle

\begin{abstract}
  \input{00_abstract.tex}
\end{abstract}

\input{01_intro}
\input{02_related_work}
\input{03_0_proposed_method}

\input{04_results}

\input{05_conclusions}

\input{06_acks}

{\small
\bibliographystyle{ieee_fullname}

\input{main.bbl}
}

\input{supplement.tex}

\end{document}

%% file: macros.tex
\usepackage{booktabs} %
\usepackage{mathtools}
\usepackage{xfrac}
\usepackage{subcaption}
\usepackage[percent]{overpic}
\usepackage{xcolor}
\usepackage{multirow}

\usepackage{colortbl}
\usepackage{siunitx}

\usepackage[ruled]{algorithm2e} %

\newcommand{\sect}[1]{Section~\ref{#1}}
\newcommand{\fig}[1]{Figure~\ref{#1}}
\newcommand{\tab}[1]{Table~\ref{#1}}
\newcommand{\eq}[1]{Equation~\ref{#1}}

\newcommand{\ourpapertitle}{Sky Optimization: Semantically aware image processing of skies in low-light photography}

\definecolor{darkgreen}{rgb}{0,0.5,0}
\definecolor{orange}{rgb}{1,0.5,0}
\definecolor{darkred}{rgb}{0.5,0,0}
\definecolor{Yellow}{rgb}{1,1, 0.6}
\definecolor{Red}{rgb}{1, 0.6, 0.6}

\newcommand{\bdarkening}{b_{\mathit{d}}}
\newcommand{\bcontrast}{b_{\mathit{c}}}
\newcommand{\tcontrast}{t_{\mathit{c}}}

\newcommand{\cundetermined}{c_{\mathit{undet}}}
\newcommand{\cdetermined}{c_{\mathit{det}}}
\newcommand{\cinpainted}{c_{\mathit{inpaint}}}

%% file: 00_abstract.tex
The sky is a major component of the appearance of a photograph, and its color and tone can strongly influence the mood of a picture. In nighttime photography, the sky can also suffer from noise and color artifacts. For this reason, there is a strong desire to process the sky in isolation from the rest of the scene to achieve an optimal look. 
In this work, we propose an automated method, which can run as a part of a camera pipeline, for creating accurate sky alpha-masks and using them to improve the appearance of the sky.
Our method performs end-to-end sky optimization in less than half a second per image on a mobile device.
We introduce a method for creating an accurate sky-mask dataset that is based on partially annotated images that are inpainted and refined by our modified weighted guided filter. We use this dataset to train a neural network for semantic sky segmentation.
Due to the compute and power constraints of mobile devices, sky segmentation is performed at a low image resolution. Our modified weighted guided filter is used for edge-aware upsampling to resize the alpha-mask to a higher resolution.
With this detailed mask we automatically apply post-processing steps to the sky in isolation, such as automatic spatially varying white-balance, brightness adjustments, contrast enhancement, and noise reduction.

%% file: 01_intro.tex
\section{Introduction}\label{sec:introduction}

Professional photographers generally invest time post-processing the appearance of the sky, as it significantly affects how humans perceive the photograph's time of day, and the relative appearance of the non-sky foreground of the scene.
Photographers will often manually segment the sky and use that segmentation to adjust the sky's brightness, contrast, color, and noise properties. This editing is particularly necessary in night-time scenes, wherein the camera receives little light and therefore produces images with significant noise. Noise in the sky can look particularly unattractive and noticeable because the sky is typically textureless.
Additionally, night-time scenes may contain a foreground that is illuminated by a nearby light source, while the sky is illuminated by scattered sunlight or by distant terrestrial lights reflected off of clouds. This means that the standard practice of using a single illuminant estimate for white balance~\cite{barron2017fast} is physically incorrect, and results in an unnatural tint of either the sky or of the foreground. This motivates our use of sky segmentation for performing spatially-varying white balance, which ameliorates this issue.

\begin{figure*}[!t]
\centering
\includegraphics[width=\textwidth]{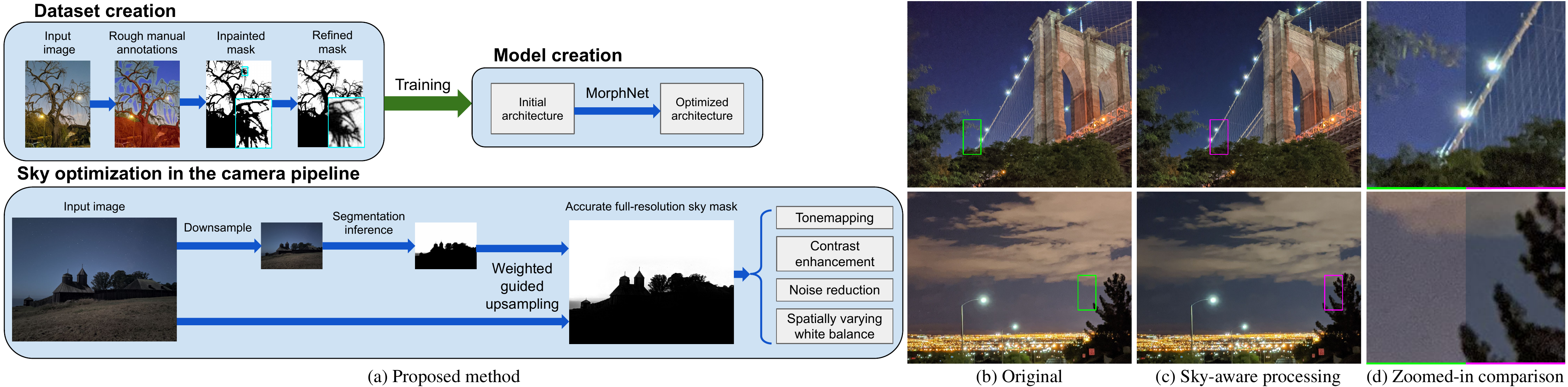}
\caption{Our method for sky-aware processing and examples. (a) We first create a dataset for sky segmentation and train a neural network to predict an alpha map of the sky. Next, we optimize the network architecture using the approach of MorphNet \cite{gordon2018morphnet}. The network is then integrated into a camera pipeline which runs inference in low resolution. Next, we run a variation of guided upsampling, to obtain a high-resolution mask. Last, we perform editing such as tonemapping, contrast enhancement, noise reduction and spatially varying auto white balance to improve the appearance of the sky and the entire image. (b) and (c) show images captured in low light, without and with sky-aware processing. (d) Zoomed-in regions of the original images (green) and sky-processed images (magenta).}
\label{fig:system_diagram}
\end{figure*}

We address the task of semantically segmenting the sky using machine learning. Because of the inherent difficulty of manually annotating a pixel-level alpha for a high-detailed sky mask, we propose an algorithmic approach for transforming approximate binary segmentation masks into accurate alpha mattes, and use this approach to improve the training data for training our model.

Our goal is to produce the sky-optimized images with a latency below one second on a mobile device with limited memory and compute resources. These constraints prohibit the use of large models, and therefore, we use the approach of MorphNet \cite{gordon2018morphnet} to shrink the model size while retaining high segmentation accuracy.

To reduce the latency and memory overhead of sky segmentation inference we reduce the size of the input image. This approach results in an output mask that is also at a reduced resolution compared to the original image. To enable high-quality image editing, this low-resolution mask is upsampled in an edge-ware way. With this high-resolution sky mask, we are able to apply various effects to the full-resolution image. Correctly performing automatic noise reduction and white balancing require an awareness of various scene properties, such as the noise level, exposure time, and illumination color, which we address in this paper.

In this work, we propose a method for semantic sky segmentation and editing that can be done on mobile devices as part of the camera pipeline. The overall framework of our approach is illustrated in \fig{fig:system_diagram}. Our contributions are:
\begin{enumerate}
    \item A method for refining the coarse sky segmentation masks of existing datasets into accurate alpha masks. We performed a user study to evaluate the quality of the refined masks and show quantitative results on the ADE20K dataset~\cite{zhou2017scene}.
    \item A method for creating accurate sky alpha masks at high resolution, using a neural network and weighted guided upsampling.
    \item A series of sky editing effects that can be achieved using the high-resolution sky masks: sky darkening, contrast enhancement, noise reduction, and auto white balance, which automatically improve the appearance of photographs, especially those captured in low light.
    \item A modification to Fast Fourier Color Constancy (FFCC) \cite{barron2017fast} to achieve spatially varying auto white balance.
    \item A complete computational photography system for performing sky segmentation and optimization on a mobile device in under half a second.
\end{enumerate}

%% file: 02_related_work.tex
\section{Related Work}
In this section, we review prior work on sky-aware image processing. Prior work on mask refinement strategies is discussed in \sect{sec:wgf}. 

Tao \etal \cite{tao2009skyfinder} created an interactive search system for skies with certain attributes and demonstrated its use for editing the appearance of skies. Owing to an interest in automatically segmenting the skies for a wide variety of applications, Mihail \etal \cite{mihail2016sky} created a dataset for evaluating sky segmentation in various conditions, and reviewed methods for sky segmentation.
Place \etal \cite{la2019segmenting} built on their prior work and evaluated the performance of a neural network, RefineNet \cite{lin2016refinenet}, for the segmentation task.
Tsai \etal \cite{tsai2016sky} used a refined sky mask to replace the sky in a photograph to give it a more interesting appearance. They utilize a fully convolutional network \cite{long2015fully} for scene parsing, followed by sky refinement with a two-class conditional random field.
In addition to segmenting and replacing the skies, Tsai \etal created a method for automatically selecting suitable sky-replacement candidates based on semantic features, and render these skies onto the target image using a per-pixel transfer process.
Unlike that work, we are not aiming to completely replace the appearance of the sky, instead, we are interested in improving the appearance of the image while producing an image that is closer to what was perceived during capture.
This is achieved by applying spatially varying white balance, noise reduction and tonemapping.
The method described by \cite{tsai2016sky} was not optimized for processing on mobile devices, for example, scene parsing alone takes 12 seconds on a desktop GPU, however \cite{halperin2019clear} proposed a sky replacement algorithm for video which can achieve nearly real-time performance on mobile devices. \cite{halperin2019clear} use a relatively small segmentation network and do not incorporate a refinement step in their performance measurements. Their results are visually pleasing for video, but are not shown for higher resolution (12 megapixel) photographs.

%% file: 03_0_proposed_method.tex
\section{Proposed Method}

\input{03_1_mask_refinement}
\input{03_2_dataset_creation}
\input{03_3_model_architecture}
\input{03_4_sky_effects}

%% file: 03_1_mask_refinement.tex
\subsection{Mask refinement}
\label{sec:wgf}

\begin{figure}[b!]
\includegraphics[width=\linewidth]{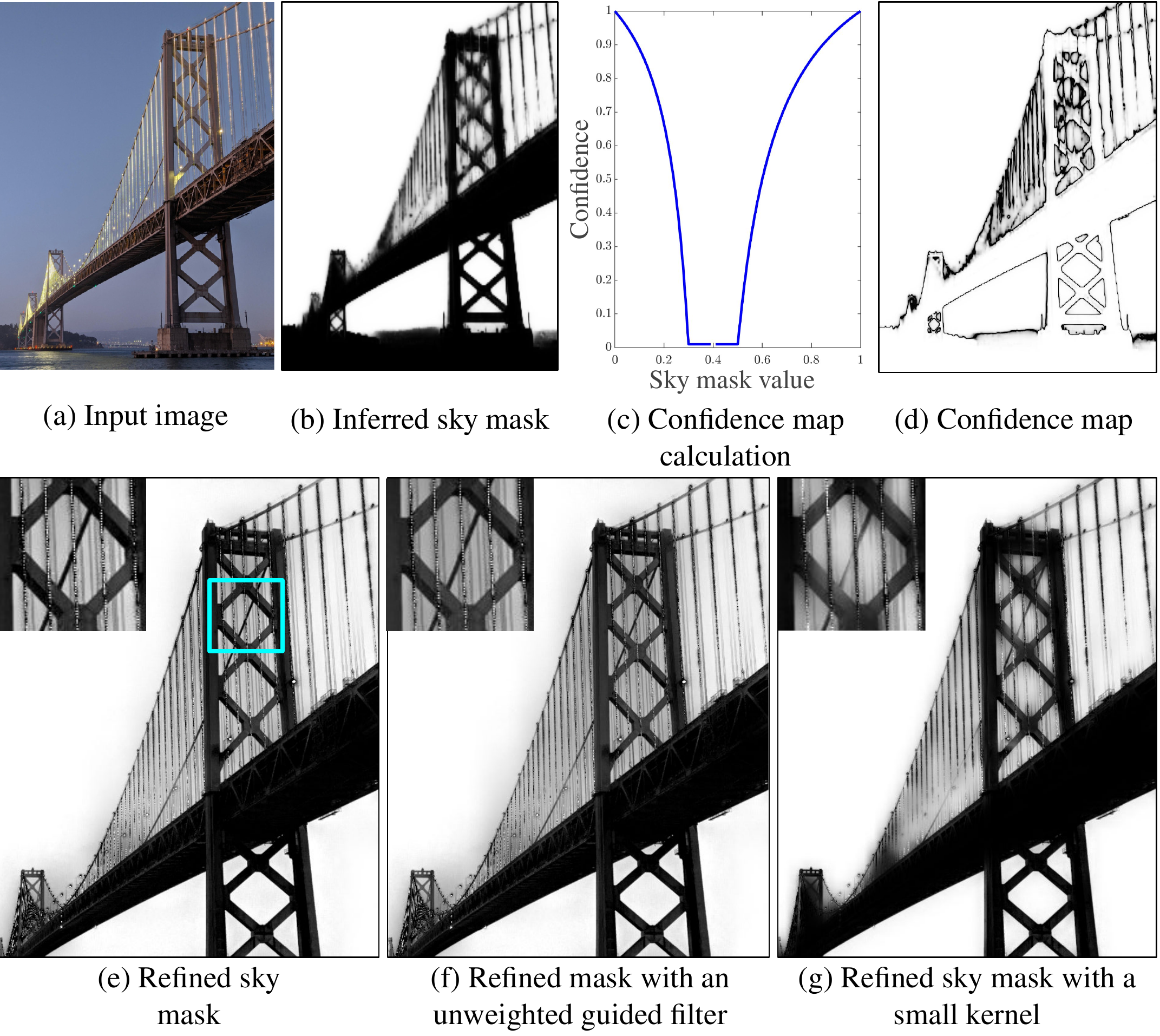}
\caption{The input to our mask refinement process is an input RGB image (a) and a coarse binary sky mask (b), shown here with nearest-neighbor upsampling. Applying our confidence map function (c) to the mask values produces a per-pixel confidence (d), which is used as input to our modified guided filter (with $s=64$) to produce the edge-aware alpha matte in (e). Using a non-weighted guided filter produces mattes with less separation between the foreground and the sky (f), and using a small spatial kernel ($s=16$) produces an inaccurate mask (g), for example, the mask does not accurately segment the cables of the bridge.}
\label{fig:wgf}.
\end{figure}

In this section we describe a guided-filter-based mask refinement method, which is used twice by our approach: First, for creating an accurate sky segmentation dataset, without tedious pixel-level manual annotations, and second, for refining the inferred low resolution sky mask, prior to using it for photo editing. An outline of this procedure can be seen in Figure~\ref{fig:wgf}.

\paragraph{The modified guided filter}
\label{sec:modified_guided_filter}

\newcommand{\ldlsolve}[2]{\operatorname{solve\_image\_ldl3}(#1, #2)}
\newcommand{\outerprod}[2]{#1 \otimes #2}
\newcommand{\refim}{I}
\newcommand{\inputim}{P} 
\newcommand{\confidence}{C}
\newcommand{\downsamplefactor}{s}
\newcommand{\wdownsample}[1]{\operatorname{weighted\_downsample}\left(#1, C, \downsamplefactor\right)} %
\newcommand{\upsample}[1]{\operatorname{smooth\_upsample}\left(#1, \downsamplefactor\right)} %
\newcommand{\downsampled}[1]{#1_{\downarrow}}
\newcommand{\upsampled}[1]{#1_{\downarrow\uparrow}}
\newcommand{\lumasmooth}{\epsilon_{\ell}}
\newcommand{\chromasmooth}{\epsilon_{c}}
\newcommand{\covref}{\downsampled{\Sigma}}
\newcommand{\covrefinput}{\downsampled{\sigma}}
\newcommand{\downA}{\downsampled{A}}
\newcommand{\downB}{\downsampled{b}}
\newcommand{\updownA}{A}
\newcommand{\updownB}{b}
\newcommand{\outputim}{Y}

The guided filter~\cite{he2010guided} is a classic technique for edge-aware filtering, which generally exhibits improved performance and speed compared to the related techniques of anisotropic~\cite{Perona90scale} or bilateral~\cite{Tomasi1998, paris2009bilateral} filtering and joint bilateral upsampling \cite{kopf2007joint}.
The guided filter has been shown to be a closed-form partial solution to the matting Laplacian~\cite{levin2007closed}, which motivates our use of it here for producing alpha mattes from sky masks.
The guided filter works by approximating the input image to be filtered (in our case, an alpha mask) as a locally affine function of the reference image (in our case, an RGB image). This is done by solving a linear least squares problem at each pixel in the image. 
In this work we present several modifications to the guided filter, which have several advantages over the traditional formulation, as listed below. The full algorithm appears in \sect{sec:modified_guided_filter}.
\begin{compactenum}
\item Our filter takes as input a per-pixel confidence map, and accordingly solves a \emph{weighted} least squares system at every pixel. This allows our filter to process masks where some values are missing, which is critical for creating high resolution alpha-masks efficiently.
\item We use a (weighted) bilinear downsample to compute the local expectations that are required for the guided filter computation and a smooth upsampling procedure to produce smooth artifact-free outputs (see \sect{sec:modified_guided_filter}) for more details).
\item Owing to our use of downsampling, the linear solver at the core of the guided filter, only needs to be evaluated on $n/s^2$ pixels, where $n$ is the number of pixels and $s$ is our downsampling factor. This approach resembles the "fast guided filter" \cite{he2010guided}.
\item We use an LDL-decomposition based solver to solve the set of linear systems. This produces more stable outputs than the conventional approach of explicitly inverting a matrix, and is faster than the approach of invoking a direct linear solver.
\item We parameterize the regularization in the (weighted) least squares solver within our filter as independent smoothness terms on the luma and chroma of the reference image, which gives us finer control over the output of the filter. 
\end{compactenum}

\paragraph{Confidence calibration}
\label{sec:confidence_calibration}
The choice of the confidence map used by our modified guided filter is critical to the quality of our edge-aware smooth sky masks.
We use different approaches for computing this confidence for the two uses of mask refinement: one for pre-processing our training data, and one for post-processing the output of our neural network.
When pre-processing our training data, we use a confidence trimap, where a value of $\cdetermined$ indicates certainty and is used for pixels which were manually annotated. $\cinpainted$  is for pixels inpainted as skies by the density estimation algorithm (described in \sect{sec:dataset_annotations}). $\cundetermined$ indicates uncertainty of the mask values and is assigned to pixels which were not annotated nor inpainted.

When post-processing the inferred sky mask from our neural network within our camera pipeline, we must adopt a different approach, as we do not have ground-truth user annotations.
Our model emits a continuous per-pixel probability, where low values correspond to ``not sky'', high values (closer to $1$) correspond to ``sky'', and intermediate values indicate uncertainty.
Accordingly, we define our per-pixel confidence measure $C_i$ a as a function of each per-pixel probability $p_i$:
\begin{align}
C_i &= \left\{
		\begin{array}{lll}
			\max\left(\epsilon, \operatorname{bias}\left(\frac{\ell - p_i}{\ell}, b \right) \right) & \mbox{if } p_i < \ell  \\
			\max\left(\epsilon,\operatorname{bias}\left(\frac{p_i - h}{1 - h}, b\right) \right) & \mbox{if } p_i > h \\
			\epsilon & \mbox{o.w.}
		\end{array}
	\right. \\
\operatorname{bias}(x; b) &= \frac{x}{(\sfrac{1}{b} - 2)(1 - x) + 1} \label{eq:bias}
\end{align}
Where $\ell=0.3$, $h=0.5$, $b=0.8$, $\epsilon=0.01$, and $\operatorname{bias}(\cdot)$ is Schlick's bias curve \cite{Schlick:1994:FAP:180895.180931}.
This function is visualized in \fig{fig:wgf}c.
This confidence is not symmetric with respect to  $p_i=0.5$, and therefore encodes a preference towards the sky label, which is done so as to ensure that sky pixels are not ignored by our filter.

\paragraph{Choosing the guided filter downsampling factor}
The choice of the downsampling factor used by the filter has a significant impact on both the guided filter's speed and spatial support.
The downsampling factor used when creating our training dataset is relatively small ($s=8$), because the image is annotated at full resolution and the initial annotations and subsequent inpainting produce good predictions except around rough edges.
In this case, the guided filter just serves to smooth the boundaries of the mask according to the reference image.
Because this step is performed before training and not on-device, the slower speed caused by this small spatial support is not problematic.
However, when we apply our guided filter variant to the output of our neural network within our camera pipeline, we use a larger downsampling factor ($s=64$). This significantly accelerates inference, and also results in improved mask quality: Because the sky mask is inferred at low resolution, small regions of the sky may not be detected as such (e.g.\, in between leaves of trees). To obtain correct sky-mask values in these regions, we need the spatial support of the filter to be large enough to allow the signal from these correctly detected sky regions to propagate to those regions where detection failed. As shown in \fig{fig:wgf}g, using a small spatial kernel negatively affects the quality of our output.

%% file: 03_2_dataset_creation.tex
\subsection{Dataset creation}
\label{sec:dataset}

Creating an accurate sky segmentation dataset is challenging because semantic segmentation datasets rely on manual annotations provided by humans. The ground-truth masks in common segmentation datasets are often coarse and inaccurate: the shape of the sky may be poorly approximated by a small number of control points, or may omit ``holes'' in other objects through which the sky is visible.
As a result, semantic segmentation models trained on these datasets do not produce accurate enough outputs to be used for our image enhancement task. 
Indeed, even using the ground-truth sky mask provided by these datasets is generally not sufficient for our task, as we will demonstrate.
It is unrealistic to rely on human annotators to produce the high degree of quality we require for our task, especially at the scale required for training modern deep neural networks. Furthermore, in order to correctly compute the sky mask in the presence of translucent objects, we require that our training data indicate the partial transparency of the foreground. Therefore, the masks should not be a per-pixel binary mask, but should instead be continuous.

In this section, we describe how we create a diverse dataset for sky segmentation in which the annotated masks are highly detailed and include continuous alpha values. In \sect{sec:comparison_dataset_refined_not_refined}, we show how dataset refinement improves the results of the segmentation model. All of the results in the paper were produced by a model trained on our dataset, unless mentioned otherwise.

\paragraph{Obtaining images for the dataset}

The images in our dataset were independently collected and include a variety of scenes, with different times of day (daytime, nighttime, sunrise, sunset), different weather conditions (clear, cloudy, foggy), challenging compositions (the sky can be seen behind trees, bridges, sculptures), reflective objects (reflective buildings and water), and astrophotography images. We have found that when a sky segmentation model is trained mostly on images with skies, it sometimes incorrectly classifies other uniform objects as being sky, such as indoor walls. We therefore used Google Images to mine for images which are similar to our false positives and added them to the training dataset. In total, our dataset includes $\sim$120,000 images.

\begin{figure}
  \centering
  \includegraphics[width=\linewidth]{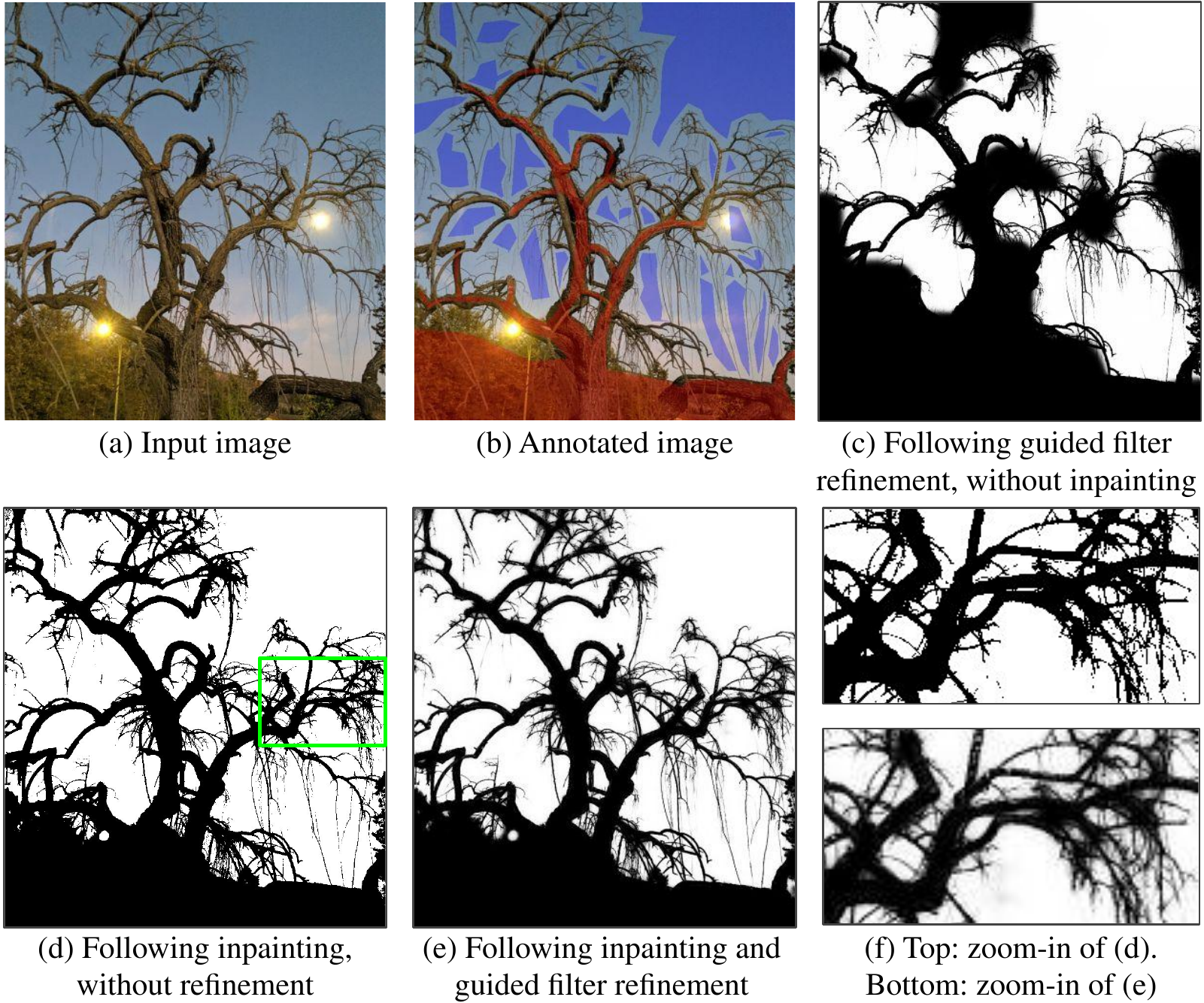}
  \captionsetup{justification=raggedright,singlelinecheck=false}
  \caption{Our method for annotating the sky segmentation dataset. (a) An RGB input image. (b) Rough partial annotations, created manually. Blue and red represent the ``sky'' and the ``not sky'' sections, respectively, and the ``undetermined'' section is not marked. (c) The manual annotations of (b) following the weighted guided filter refinement process without inpainting. In this case all the undetermined pixels are labeled as ``not sky'' and are given confidence values of $\cundetermined=0.5$. (d) The manual annotations following density estimation inpainting. The mask is binary. (e) The mask of (d) following refinement with the weighted guided filter. This mask is detailed and includes continuous values, as can be seen in the zoomed-in crops in (f).}
  \label{fig:de}
\end{figure}

\paragraph{Refined and Inpainted Annotations}
\label{sec:dataset_annotations}
Here we show a method for efficiently creating an accurate continuous-value alpha mask of the sky. We start with coarse manual annotations dividing the image into three sections: ``sky'', ``not sky'' and ``undetermined''. The annotations were made by human annotators who clicked to create polygons of the three different sections. The ``sky'' section includes skies and objects in the skies, such as clouds, the moon and stars. The ``not sky'' section was annotated such that it does not include any skies or partially transparent foregrounds. The mined negative images of uniform sky-like surfaces are of indoor scenes and are therefore globally annotated as ``not sky''. The ``undetermined'' label is used for any pixel that is not annotated by the previous two labels, such as boundaries between trees and the sky (\fig{fig:de}b). This region may contain both ``sky'' and ``not sky'' pixels that would be impractical to manually annotate, or that contain transparency and therefore require a non-binary alpha value. These ``undetermined'' areas are inpainted by using density estimation \cite{parzen1962estimation}, as described below. 

Density estimation uses the distribution of RGB values of the annotated ``sky'' pixels to inpaint the pixels in the ``undetermined'' section. Here we take advantage of the limited variability of RGB values in the skies. We calculate the probability that each ``undetermined'' pixel belongs to the ``sky'' pixels using \eq{eq:de}.
All ``undetermined'' pixels with a probability $p_i$ greater than a threshold $p_c=0.6$ are re-labeled as being ``sky'', while those with probabilities below that threshold are re-labeled as ``not sky'' and assigned a low confidence of $\cundetermined$ (as described in \sect{sec:confidence_calibration}). We then apply our modified weighted guided filter to the mask.

%% file: 03_3_model_architecture.tex
\subsection{Model architecture}
\label{sec:fluidnets}

At the core of our model is a UNet~\cite{ronneberger2015unet} neural network that takes as input an RGB image and predicts as output the per-pixel likelihood that each pixel is the sky. The specific model we build upon is a three-stage UNet model that was previously used for portrait segmentation on mobile devices \cite{wadhwa2018synthetic}. 

To optimize the performance of the network and reduce its size we used the MorphNet method~\cite{gordon2018morphnet}. MorphNet balances the model's loss and its resource constraints, in terms of model size, to discover architectures that are more efficient. It does that by relaxing the discrete model parameterization into a continuous formulation that allows for the model structure to be learned jointly along with its weights during training. We follow the previously-described 2-step approach of first shrinking and then expanding the model. Our criteria during  meta-optimization was to improve the evaluation metric while reducing the size of the model. Following this optimization step, we obtain a model that is more accurate (the ${IoU}_{0.5}$ increases from 0.9186 to 0.9237) and 27\% smaller (see \sect{sec:model_optimization} for more detail). Next, we quantize all model weights from 32-bit to 16-bit floating point, which further reduces the model size by half with a negligible quality gap. The final size of our model is 3.7 MB.

%% file: 03_4_sky_effects.tex
\subsection{Sky optimization}
\label{sec:sky_optimization}

Low light photography presents specific challenges which, if not addressed, degrades the appearance of the skies. Specifically, low light images often have a low signal-to-noise ratio, multiple illumination sources, and a high dynamic range, which may cause the skies to appear too noisy and have an unnatural color. Additionally, a side-effect of brightening the foreground in low-light photographs is over brightening the skies and creating a night-into-day effect. Therefore, it is beneficial to separately tune tone, noise reduction and color for the skies, which ultimately improves the image quality of the entire photograph.

\begin{figure}
\includegraphics[width=\linewidth]{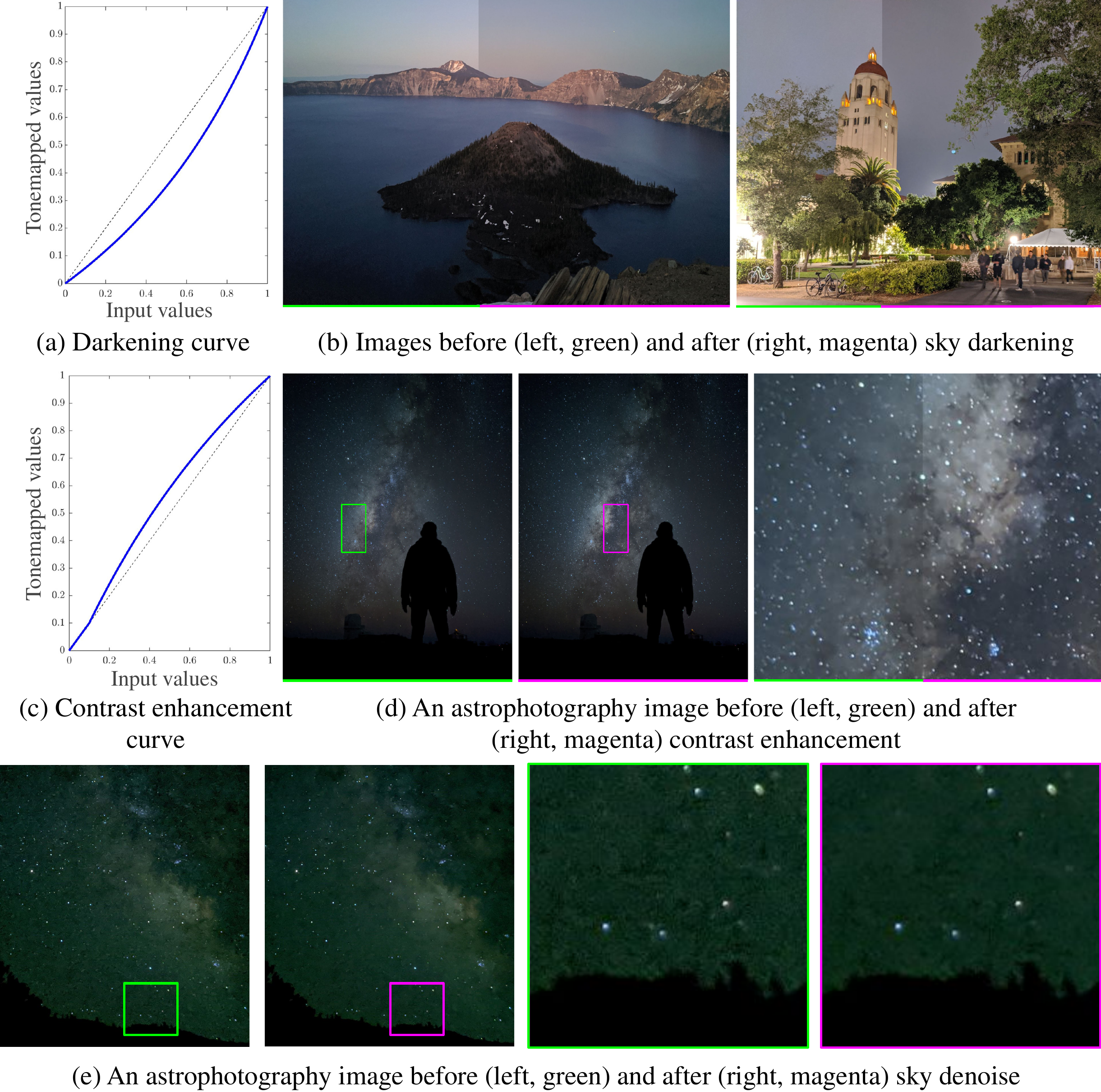}
\caption{Sky tonemapping and denoise. (a) The tonemapping curve we used to darken the sky to reduce the night-into-day effect of low-light imaging. The curvature is determined by a $\bdarkening$ which we calibrate according to the brightness of the sky and the scene. (b) Examples of original (left) and sky-darkened images (right). Notice that only the sky brightness is changed while the foreground remains the same. (c) The tonemapping curve used for contrast enhancement. (d) A photograph of the Milky Way before (left) and after sky contrast enhancement (right). (e) The sky mask allows us to tune the denoising algorithm to improve the appearance of the sky without affecting details in the foreground. The mid-frequency noise blotches are removed while the high resolution details of the stars are retained. Readers are encouraged to zoom-in to see the difference in noise characteristics.}
\label{fig:sky_effects}
\end{figure}

\paragraph{Sky darkening and contrast enhancement}
Tonemapping is a technique used in image processing to map one set of colors or brightness values to another. We use a tonemapping curve to darken the skies in low light images. We apply tonemapping on the V channel following HSV decomposition~\cite{smith1978color}. The shape of our tonemapping curve is a bias curve, $\operatorname{bias}(v;\bdarkening)$, as shown in \eq{eq:bias} and \fig{fig:sky_effects}a. In $\operatorname{bias}(v;\bdarkening)$, $v$ is the pixel value, which after the HSV decomposition is in a range of $[0,1]$, and $\bdarkening$ is a parameter that controls the amount of darkening. When $\bdarkening=0.5$, the output of the tonemapping curve is identical to its input (dashed line in \fig{fig:sky_effects}a). As $\bdarkening$ decreases, the output decreases in the mid-range, meaning that the skies become darker. In order to automatically choose the ``right'' tone of the sky, the parameter $\bdarkening$, which indicates the amount in which the sky will be darkened, is calibrated per image as a function of the sky brightness and the brightness of the scene. The ``right'' amount of darkening is highly subjective, as we are aiming to reproduce the \textit{feel} of the scene as it was captured. To achieve this, we tagged hundreds of photos at various conditions. We binned and averaged these tags to learn a 2-dimensional look-up table that maps the brightness of the scene and the sky to the amount of sky darkening, as represented by $\bdarkening$.

A subsequent tonemapping curve is used to enhance the contrast of images with certain characteristics. Specifically, contrast enhancement is targeted towards astrophotography and is designed to boost the appearance of stars and other celestial objects, such as the Milky Way. The contrast enhancement curve, shown in \fig{fig:sky_effects}c and \eq{eq:contrast}, also takes as input the pixel value $v_i$. It leaves the very low-brightness pixels unchanged and uses a bias curve to enhance the contrast of the mid-brightness pixels:
\begin{align}
v_o &= \left\{
		\begin{array}{lll}
			v_i  & \mbox{if } v_i < \tcontrast   \\
			(1-\tcontrast)\operatorname{bias}\left(\frac{v_i - \tcontrast}{1-\tcontrast};\bcontrast\right) + \tcontrast & \mbox{if } v_i \geq \tcontrast \\
		\end{array}
	\right.
\label{eq:contrast}
\end{align}
The parameters $\bcontrast$ and $\tcontrast$ indicate the intensity and the range of contrast enhancement. We calibrate $\bcontrast$ as a function of the exposure time and the noise in the image, in order to avoid enhancing the contrast of noise.
We labeled tens of photos and performed binning and averaging of the tags to determine a 2-dimensional mapping from the exposure time and signal-to-noise to $\bcontrast$. The threshold, $\tcontrast$, determines the range of pixels which remain unchanged and pixels which will be contrast-enhanced. We have empirically found that a value of $\tcontrast=0.085$ yields a pleasing contrast enhancement of the stars and the Milky Way. After tonemapping, the resulting pixel values and the original pixels values are alpha blended using the sky mask as the weight.

\paragraph{Sky denoising}
The appearance of the sky tends to be much more regular and predictable in terms of its image content compared to other parts of the scene. This allows us to tune a denoising algorithm for the sky that more aggressively removes noise from smooth regions, while preserving high-frequency details such as stars and mid-frequency details such as clouds. We do this by tuning the frequency response of the spatial luma denoising algorithm of \cite{liba2019handheld}, which uses a pyramid based bilateral denoise algorithm with four levels. In order to optimize noise reduction for the skies, we increase the amount of denoising at the two higher levels (low resolution noise) up to $2.5\times$. We only slightly increase the denoising strength, up to $5\%$, in the two lowest levels (highest resolution). The amount in which we increase the denoising strength in the skies is a function of the signal-to-noise ratio of the image and was calibrated by labeling tens of low-light images. We use the sky mask to indicate where to use these modified denoising parameters, and in the foreground we use the same denoising parameters as in \cite{liba2019handheld}. We combine the separately denoised foreground and sky using an adjusted sky mask as an alpha in a weighted average. In order to protect the details in the foreground, we adjust the sky mask by clamping to $0$ the mask values below a threshold, $t_d=0.8$, and scaling the remaining mask values between $0$ and $1$. \fig{fig:sky_effects}e shows an example in which noise reduction increases in the sky while the foreground and the high-frequency stars remain unchanged. Additional examples and a comparison to an end-to-end neural network \cite{chen2018learning} are in \sect{sec:sky_denoise_L2SITD}.

\begin{figure}
\includegraphics[width=\linewidth]{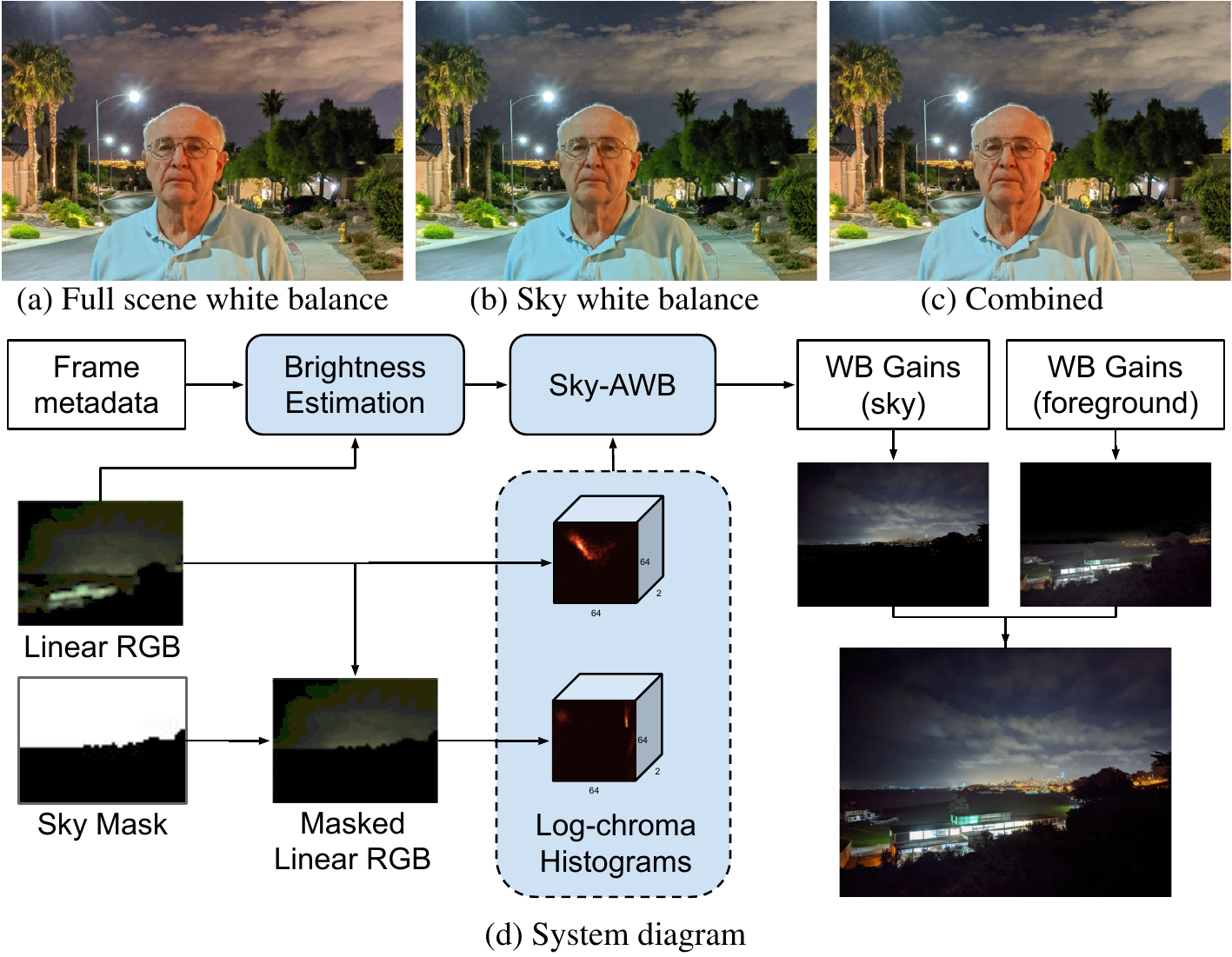}
\caption{(a-c) Conventional white balance algorithms assume a single global illuminant color, but in night-time photographs, often the sky is lit differently than the foreground. Prioritizing the foreground produces a natural looking face but a magenta sky (a), while prioritizing the sky corrects the background but makes the subject look blue (b). With sky segmentation we are able to estimate two distinct white balance gains, which are applied according to our estimated sky mask (c). (d) A system diagram of our spatially varying auto white balance system, based on FFCC \cite{barron2017fast}. Using the sky mask, we provide additional log-chroma histograms and log-brightness estimates to FFCC that isolate the color distribution of the sky. This FFCC sky model is trained separately on the dataset that optimizes the color of the sky, and we additionally train the standard model of \cite{barron2017fast} to take the entire image as input and optimize for the color of the foreground. The final rendition blends the sky and the foreground with their separate white balance gains according to our predicted sky mask to result in a spatially-varying white balance. }
\label{fig:sky_awb}
\end{figure}

\paragraph{Dual auto white balance}
A conventional image processing pipeline is often constrained by the assumption of a single global white balance: pipelines assume that all pixels are lit by one single illuminant, and attempt to estimate and remove that one illuminant.
This can produce suboptimal results in scenes with mixed illuminants, which is often the case for images of the night sky, which is far in the background and is often illuminated differently from the foreground.
We observe that conventional global white balance algorithms often prioritize the color fidelity of the foreground over background, which can compromise image quality for the sky region. \fig{fig:sky_awb}a demonstrates the global white balance of \cite{barron2017fast,liba2019handheld}, which  prioritizes the color of the person over the color of the sky, thereby causing the sky to look magenta. If this global white balance algorithm were to instead produce the gains that correct for the color of the sky and apply those gains globally, as shown in \fig{fig:sky_awb}b, the subject's skin would appear unnaturally blue. By using sky segmentation, we are able to produce two distinct white balance estimates, which we can use to separately correct the colors of the foreground and of the sky, as shown in \fig{fig:sky_awb}c.

Our dual white balance algorithm is built upon the Fast Fourier Color Constancy (FFCC) model \cite{barron2017fast} using a modified loss function for low-light images \cite{liba2019handheld}. Unlike past work, we calculate two distinct auto white balance (AWB) gains: one for the foreground, and one for the background, using the sky mask.
This is done by modifying FFCC to allow it to reason about the chroma distribution of the sky independently from the foreground during training (\fig{fig:sky_awb}d), which results in two models: one for the entire scene and another just for the sky. These two models were trained on two datasets, one that was tagged to optimize the colors of the entire scene, prioritizing people and foreground objects, and another that was tagged to optimize only the color of the sky. The input for training the former model is the entire image, while the input for training the sky model includes both the entire image and a version of the image in which only the sky is visible. These two models, and the sky mask, are then used in the camera pipeline to separately calculate the white balance gains for the entire scene and for the sky. We apply the white balance gains to the image using the sky mask estimated by our model as the alpha in a weighted average, resulting in a final composition where the foreground and sky have been independently white balanced.

%% file: 04_results.tex
\section{Experimental Results}
\label{sec:results}
\label{sec:comparison_dataset_refined_not_refined}
Here we show how the refinement of the annotated masks in the training dataset improves the sky masks inferred by the segmentation model, and that the quality of the mask can be further improved by weighted guided upsampling. In order to do this, we first establish, with a user study, that non-refined annotations can be qualitatively improved by refinement, and specifically by the process that we developed and describe in \sect{sec:dataset}. These results, which generally demonstrate the significance of dataset refinement, are shown on a public dataset, ADE20K \cite{zhou2017scene}, and a baseline UNet model, for the purpose of reproducibility. 

\begin{table*}
  \centering
  \resizebox{\linewidth}{!}{
  \begin{tabular}{rl|ccccccccc}
                 Training Data & Algorithm & $\operatorname{mIOU_{0.5}}\uparrow$ & $\operatorname{BL}\downarrow$ & $\operatorname{MCR_{0.5}}\downarrow$ &
                 $\operatorname{RMSE}\downarrow$ &
                 $\operatorname{MAE}\downarrow$ &
                 $\operatorname{JSD}\downarrow$ \\
\hline
  ADE20K & UNet + Bilinear Upsampling             & 0.926          & 0.0530          & 0.0151          & 0.0986          & 0.0174          & .00753 \\
  ADE20K+GF & UNet + Bilinear Upsampling          & 0.920          & 0.0511          & 0.0162          & 0.1051          & 0.0186          & .00843 \\
  ADE20K+DE+GF & UNet + Bilinear Upsampling       & \textbf{0.936} & 0.0498          & \textbf{0.0131} & \textbf{0.0910} & \textbf{0.0154} & \textbf{.00645} \\
   ADE20K & UNet + Guided Filter Upsampling             & 0.933          & 0.0476          & 0.0137          & 0.0999          & 0.0258          & .00993 \\
  ADE20K+GF & UNet + Guided Filter Upsampling          & 0.922          & 0.0482          & 0.0159          & 0.1081          & 0.0270          & .01092 \\
  ADE20K+DE+GF & UNet + Guided Filter Upsampling  & 0.935          & \textbf{0.0465} & 0.0134          & 0.0972          & 0.0250          & .00948 \\
  \end{tabular}
  }
  \captionsetup{justification=raggedright,singlelinecheck=false}
  \caption{Evaluation of the various models with ADE20K+DE+GF as the ground-truth.}
  \label{tab:results_model_comparison}
\end{table*}

\subsection{Establishing a refined sky-segmentation dataset}

The ADE20K dataset \cite{zhou2017scene} has binary mask annotations for various labels, including ``sky''. We refine this dataset with two different methods: 1) using the guided filter only (ADE20K+GF), and 2) using density estimation inpainting and the guided filter (ADE20K+DE+GF). Additional details, such as the steps for creating these datasets and example images, are in \sect{sec:dataset_refinement}.

To empirically determine which sky masks are more accurate, we conducted a user study in which the users were asked to choose their preferred mask from a pair of masks from either the raw ADE20k dataset, ADE20K+GF or ADE20K+DE+GF. To show the accuracy of the masks, we apply a sky darkening algorithm that blends a black image with the original image according to the sky mask. The 21 participants of the study preferred the masks of ADE20K+DE+GF in 82.6\% of the cases, when compared to raw ADE20K (more details on the results of the user study are in \sect{sec:user_study_results}). Therefore, we conclude that the masks refined by our DE+GF algorithm are more accurate than the raw ADE20K annotations, and we will evaluate sky segmentation using the refined masks as the ground truth in the next section. 
Note that current public sky segmentation datasets consist of binary masks. As shown by our user study, binary masks produce worse image quality when used for sky-aware effects. Therefore, the results of an evaluation on these binary datasets would not be a meaningful indicator for our intended task.

\subsection{Model evaluation}
After establishing that the ADE20K+DE+GF dataset is more accurate compared to the raw annotations, we proceed by evaluating models trained on differently refined datasets on this new ground truth. We trained three neural networks, all with a three-stage UNet architecture, on 1) the raw ADE20K dataset, 2) the ADE20K+DE+GF dataset described above, and 3) the ADE20K dataset refined using the guided filter without inpainting (ADE20K+GF). To mimic the processing of the camera pipeline, we downsample the input images and ground truth masks during training to a resolution of $256\times256$. Evaluation is performed at full resolution, after upsampling by either a bilinear algorithm or our weighted guided filter algorithm.

\begin{figure}
\includegraphics[width=\linewidth]{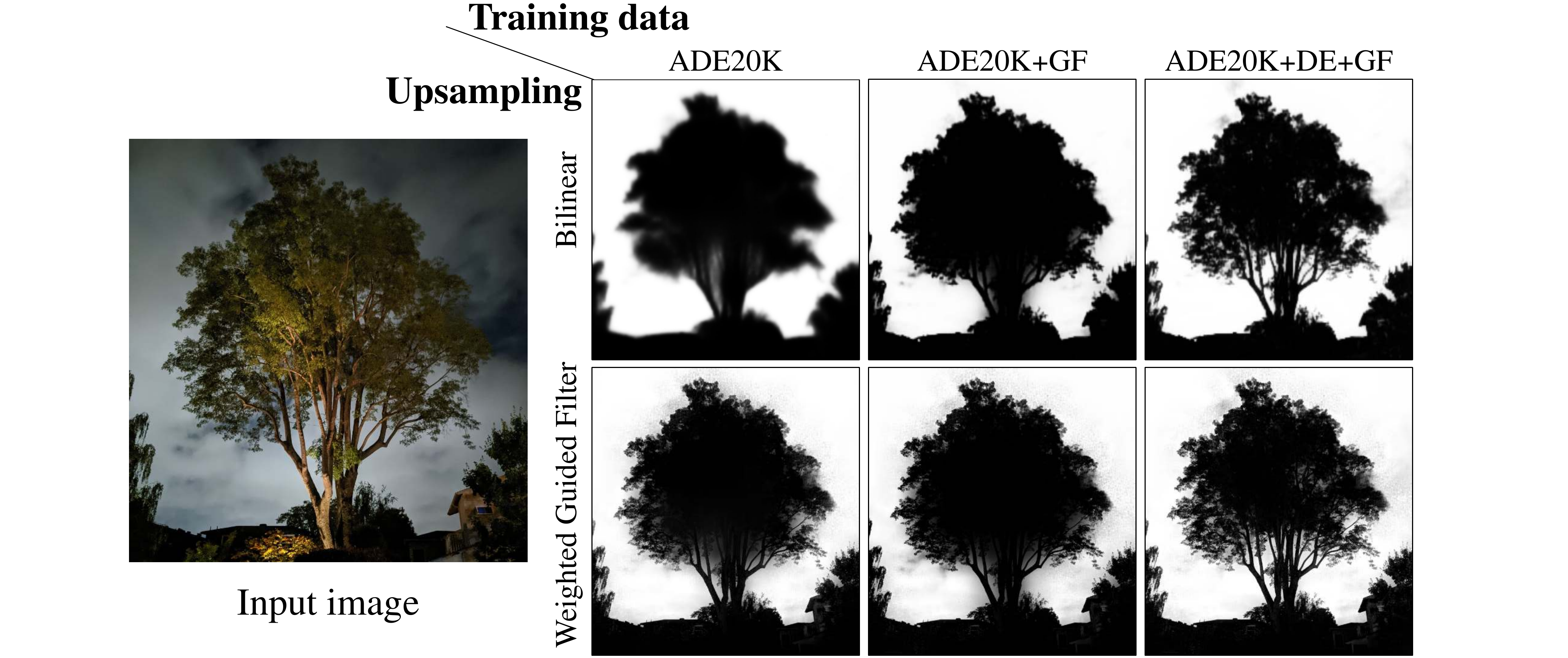}
\caption{Sky masks of the image shown on the left, computed with models that were trained on different datasets: raw ADE20K, ADE20K+GF, and ADE20K+DE+GF. The masks are inferred at low resolution ($256\times256$) and then upsampled by either bilinear upsampling or our modified weighted guided filter upsampling. This example demonstrates that the masks produced by the model that was trained on the refined ADE20K+DE+GF are more accurate compared to the other masks, even after guided filter upsampling. This observation correlates well with \tab{tab:results_model_comparison}. We observe that guided filter upsampling produces more detailed masks compared to bilinear upsampling, even though this quality is not manifested by most of the evaluation metrics.}
\label{fig:model_comparison}
\end{figure}

The evaluation metrics, which are described in \sect{sec:evaluation_metrics}, are: the mean intersection-over-union (mIOU), misclassification rate (MCR), root mean square error (RMSE), mean absolute error (MAE), boundary loss (BL), and Jensen-Shannon Divergence (JSD). The results of the evaluation are in \tab{tab:results_model_comparison} and show that the model that was trained on the refined masks (DE+GF) performs better compared to the models trained on the raw masks or the masks refined only by the guided filter (GF, without DE inpainting). Even following guided filter upsampling, training on the refined dataset is beneficial, meaning that both refinement processes are needed to produce optimal results and that guided filter upsampling cannot entirely correct imperfect masks. Perhaps surprisingly, the evaluation metrics are better in most cases for the bilinear upsampled masks. This could be attributed to the fact that upsampling and refinement are not a part of our training pipeline. That said, when looking at the full resolution images, the masks upsampled using the guided filter have more detail and seem to better represent the image (\fig{fig:model_comparison}). Importantly, guided filter upsampling performs better on the BL metric, therefore, it appears that the BL metric correlates with perceptual quality better than the other metrics.

\subsection{Implementation details and performance}
We integrated sky segmentation and processing into a camera pipeline which is implemented on Android and relies on the Camera2 API. The sky mask is inferred on an RGB image which was downsampled to a resolution of $256\times256$. The computation of the segmentation mask is implemented on a mobile GPU and requires ~50 ms. The inferred mask is upsampled using our modified weighted guided filter to a quarter resolution of the image ($1024\times768$). This step was implemented in Halide \cite{ragankelley:halide:2012} and has a latency of 190 ms. Tonemapping and the white balance gains are applied at the quarter resolution and have a latency of ~47 ms. The mask is bilinearly upsampled to full resolution ($4032\times3024$) for sky denoising. The latency for bilinear upsampling and image composition, used for denoising, is 160 ms. Example results are shown in \fig{fig:system_diagram} and \sect{sec:additional_examples}.

%% file: 05_conclusions.tex
\section{Conclusions}
\label{sec:conclusions}
We have presented a method for creating accurate sky masks and using them to adjust the appearance of the sky. 
We have shown that refinement is beneficial for creating a dataset for training a segmentation model and for upsampling the mask following inference by that model. The sky mask enables us to edit the skies within the camera pipeline in under half of a second on a mobile device. The edits are particularly beneficial in low light imaging, when the color of the sky is affected by the global white balance and when noise is significant. A variation of our sky optimization system was launched on Google Pixel smartphones as part of “Night Sight”.

%% file: 06_acks.tex
\section{Acknowledgements}
\label{sec:acknowledgements}
The algorithms presented in this paper are a result of an ongoing collaboration between several teams in Google Research. From Gcam we particularly thank Marc Levoy, Kiran Murthy, Michael Milne, and Andrew Radin; from the Mobile-Vision team, we thank Emily Manoogian, Nicole Maffeo, Tomer Meron, Weijun Wang and Andrew Howard; from the Luma team, Sungjoon Choi.

%% file: supplement.tex
\appendix

\section{The modified guided filter algorithm}
\label{sec:modified_guided_filter}
In this section we describe our modified guided-filter-based mask refinement method. Pseudo code for our modified guided filter is as follows:
\begin{align}
\hline
\operatorname{modified\_guided\_filter}(\refim, \inputim, \confidence, \downsamplefactor, \lumasmooth, \chromasmooth): \span \quad\quad\quad\quad\quad\quad\quad\quad  \nonumber\\
\quad\downsampled{\refim} &= \wdownsample{\refim} \nonumber\\
\quad\downsampled{\inputim} &= \wdownsample{\inputim} \nonumber\\
\quad\covref &= \wdownsample{\outerprod{\refim}{\refim}} - \outerprod{\downsampled{\refim}}{\downsampled{\refim}} \nonumber\\
\quad\covrefinput &= \wdownsample{\refim \circ \inputim} - \downsampled{\refim} \circ  \downsampled{\inputim} \nonumber\\
\quad\covref &= \covref + \begin{bmatrix}
\lumasmooth^2 &  &  \\
 & \chromasmooth^2 &  \\
 &  & \chromasmooth^2 \\
\end{bmatrix} \nonumber\\
\quad\downA &= \ldlsolve{\covref}{\covrefinput} \nonumber\\
\quad\downB &= \downsampled{\inputim} - \downA \cdot \downsampled{\refim} \nonumber\\
\quad\updownA &= \upsample{\downA} \nonumber\\
\quad\updownB &= \upsample{\downB} \nonumber\\
\quad\outputim &= \updownA \cdot \refim + \updownB \\
\hline \nonumber
\end{align}
Inputs to the filter are: a 3-channel reference image $\refim$ (assumed to be in YUV),
the quantities to be filtered, $\inputim$ (in our case, the sky mask), a confidence map $\confidence$,
and hyperparameters: $\downsamplefactor$, the downsampling factor, and $\lumasmooth$, and $\chromasmooth$, the regularization factors for the luma and chroma, respectively.

The output of the filter is $\outputim$, a mask that resembles $\inputim$ where $\confidence$ is large, and adheres to the edges in $\refim$.
Regarding notation, $\circ$ is the Hadamard product (where 1-channel images are ``broadcasted'' to match the dimensions of images with more channels), and $\cdot$ is a dot product (Hadamard product that is then summed over channels).
The outer product of two images $A = \outerprod{X}{Y}$ is defined as taking two 3-channel images $X$ and $Y$ and producing a 6-channel image $A$ representing the upper-triangular portion of the outer product of each pixel of $X$ and $Y$:
\begin{align}
A_{1,1} = X_{1} \circ Y_{1}, \quad
A_{1,2} = X_{1} \circ Y_{2}, \quad
A_{1,3} &= X_{1} \circ Y_{3} \nonumber \\
A_{2,2} = X_{2} \circ Y_{2}, \quad
A_{2,3} &= X_{2} \circ Y_{3} \nonumber \\
A_{3,3} &= X_{3} \circ Y_{3}
\label{eq:outer3}
\end{align}
Our weighted downsample, $\operatorname{weighted\_ds}$, is simply a standard bilinear downsample operator applied using homogeneous coordinates:
\begin{equation}
\wdownsample{X} = \frac{\operatorname{ds}(X \circ C, \downsamplefactor)}{\operatorname{ds}(C, \downsamplefactor)}
\end{equation}
where division is element-wise, and $\operatorname{ds}(\cdot, \downsamplefactor)$ is bilinear downsampling according to a spatial bandwidth $\downsamplefactor$. $\upsample{\cdot}$ is the smooth upsampling procedure, described next. $\ldlsolve{A}{b}$ is an LDL-decomposition based linear solver designed to operate on 3-channel images, from \cite{Valentin2018}. For completeness, this algorithm is reproduced below.

\begin{figure}[b!]
\includegraphics[width=\linewidth]{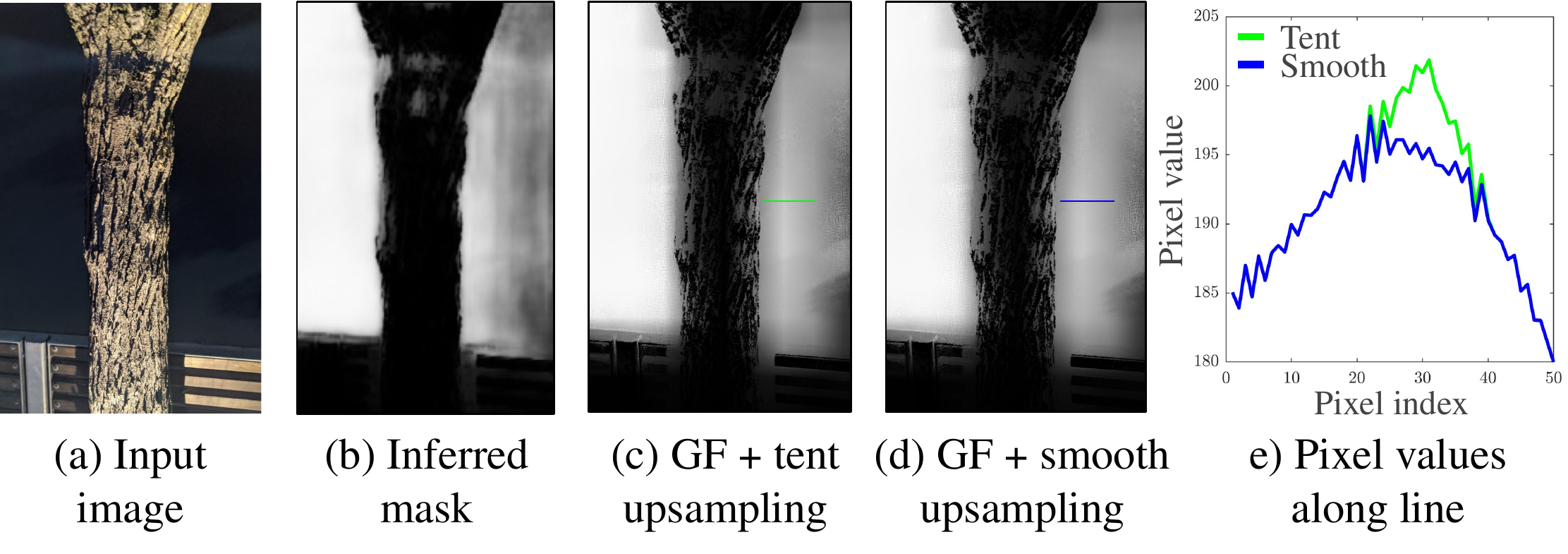}
\caption{Using bilinear upsampling within a guided filter (GF) results in noticeable triangle-shaped artifacts (c, e), while our three-step upsampling results avoids such artifacts (d, e).}
\label{fig:wgf_tent}
\end{figure}

The traditional guided filter uses a box filter to compute local expectations of various quantities. Because the box filter is applied twice, and because the convolution of two box filters is a triangle-shaped ("tent") filter, the output of the traditional guided filter tends to contain triangle-shaped artifacts. Though there exist fast techniques for applying smoother blur kernels than box filters~\cite{Young1995RecursiveIO}, these techniques are still significantly more expensive than box filters, and do not reduce the number of linear systems to be solved. 
In our algorithm, smooth upsampling is achieved by applying the triangle-shaped convolution kernels consecutively (in our case, in 3 steps), which effectively changes the shape of the interpolation kernel to be smoother, and significantly reduces upsampling artifacts. For example, for a downsampling factor $s=64$, instead of upsampling with a single kernel with a support of 64 $\times$ 64, we use triangle kernels with a support of 4 $\times$ 4 three times, one after the other. We chose this method of linear upsampling rather than a more advanced method owing to its separability and efficient implementation. The effect of this smooth upsampling can be seen in \fig{fig:wgf_tent}.

Note that our modified guided filter formulation degrades naturally to the traditional formulation of the guided filter if 1) $\wdownsample{\cdot}$ and $\upsample{\cdot}$ are both replaced with a box filter, 2) the reference image $\refim$ is RGB and $\lumasmooth=\chromasmooth$, and 3) $\ldlsolve{\cdot}{\cdot}$ is replaced with matrix inversion and a matrix multiply.

Our approach of accelerating part of the guided filter through the use of spatial downsampling is superficially similar with the ``fast guided filter'' \cite{He015}, which also yields an acceleration from $\mathcal{O}(n)$ to $\mathcal{O}(n/s^2)$ for an intermediate step of the filter. This is accomplished by simply subsampling the input mask before computing the affine filter coefficients, which are then applied to the full resolution mask. Though fast, this approach ignores the vast majority of the input mask, and thereby assumes that the input to the filter is very smooth and regular. This does not hold in our use case: for example, if we had a single high-confidence pixel surrounded by many low-confidence pixels, we would require a guarantee that this single pixel's value would propagate to all nearby low-confidence pixels in the output, and a subsampling approach will not guarantee this (and worse, will cause the output of the model to vary significantly depending on whether or not the single high-confidence pixel happens to lie at one of the few spatial locations that is subsampled). In contrast, our approach ignores none of the input mask, is completely deterministic, and still yields the same asymptotic acceleration of the filter's internal linear solver step.

\subsection{LDL-decomposition based linear solver}
In the refinement algorithm, $\ldlsolve{A}{b}$ is an LDL-decomposition based linear solver designed to operate on 3-channel images, from \cite{Valentin2018}. For completeness, we reproduce that algorithm here:
\begin{align}
\hline
\ldlsolve{A}{b}: \span\quad\quad\quad\quad\quad\quad\quad\quad\quad\quad\quad \nonumber\\
\quad d_1 &= A_{1,1}\nonumber\\
\quad L_{1,2} &= A_{1,2} / d_1\nonumber\\
\quad d_2 &= A_{2,2} - L_{1,2} \circ A_{1,2}\nonumber\\
\quad L_{1,3} &= A_{1,3} / d_1\nonumber\\
\quad L_{2,3} &= (A_{2,3} - L_{1,3} \circ A_{1,2}) / d_2\nonumber\\
\quad d_3 &= A_{3,3} - L_{1,3} \circ A_{1,3} - L_{2,3} \circ L_{2,3} \circ d_2\nonumber\\
\quad y_1 &= b_1\nonumber\\
\quad y_2 &= b_2 - L_{1,2} \circ y_1\nonumber\\
\quad y_3 &= b_3 - L_{1,3} \circ y_1 - L_{2,3} \circ y_2\nonumber\\
\quad x_3 &= y_3 / d_3\nonumber\\
\quad x_2 &= y_2 / d_2 - L_{2,3} \circ x_3\nonumber\\
\quad x_1 &= y_1 / d_1 - L_{1,2} \circ x_2 - L_{1,3} \circ x_3 \\
\hline \nonumber
\end{align}
Where the inputs to this function are a 6-channel image $A$ and a 3-channel image $b$, with channels in $A$ corresponding to the upper triangular part of a $3 \times 3$ matrix.
The output of this function is a 3-channel image $x$ where for each pixel $i$ in the input linear system, $x(i) = A(i) \backslash b(i)$ using an LDL decomposition.

\section{Density estimation algorithm}
\label{sec:density_estimation_algorithm}
The density estimation algorithm is used to inpaint unlabeled pixels and partially automate the process of annotating sky masks. The probability that an unlabeled pixel belongs to the ``sky'' pixels is described in \eq{eq:de}, in which $i$ indicates an ``undetermined'' pixel and $j$ indicates a ``sky'' pixel:
\begin{align}
    p_i &= \frac{1}{|\{ \mathrm{sky} \}|} \sum_{j \in \{ \mathrm{sky} \}}  K\left(I_i,I_j \right)
    \label{eq:de} \\
    K(I_i, I_j) &= \frac{1}{\left(2 \pi \sigma^2\right)^{3/2}} \exp\left(-{\frac {\sum_c (I_i^{c} - I_j^{c})^2}{2\sigma^2 }}\right)
\end{align}
$K(\cdot)$ is a multivariate Gaussian kernel with a Euclidean distance between the RGB values of pixels, assuming a diagonal covariance matrix ($c$ indicates the color channel). We use $\sigma=0.01$ as the kernel's standard deviation.
In practice, to reduce computation time we sample $1024$ sky pixels uniformly at random to compute these probabilities. 

\section{Model optimization}
\label{sec:model_optimization}
Table~\ref{tab:model_architecture} shows a comparison of the size, latency, and IoU scores of the original model and of the two MorphNet steps. At the end of model optimization we arrive at a model that is more accurate and 27\% smaller.

\begin{table}
  \centering
  \begin{tabular}{lccc}
                 & Original & MorphNet & MorphNet\\
                 & UNet & Shrink  & Expand\\
  \toprule
  Model Size (MB)       & 5.1 & 1.7 & 3.7 \\
  ${IoU}_{0.5}\uparrow$ & 0.9186 & 0.8895 & \textbf{0.9237} \\
  Size Reduction        &  & 66\% & 27\% \\
  \bottomrule
  \end{tabular}
  \captionsetup{justification=raggedright,singlelinecheck=false}
  \caption{The model performance before and after optimization with MorphNet and weight quantization to float-16. Inference latency is measured on a $256 \times 256$ image on a mobile CPU (Qualcomm Snapdragon 845) with $4$ threads. Evaluation is performed on our internal dataset, and therefore these results cannot be directly compared with those in Table 1 of the main paper.}
  \label{tab:model_architecture}
\end{table}

\section{Sky denoise and comparison to CNN}
\label{sec:sky_denoise_L2SITD}
\begin{figure}
\includegraphics[width=\linewidth]{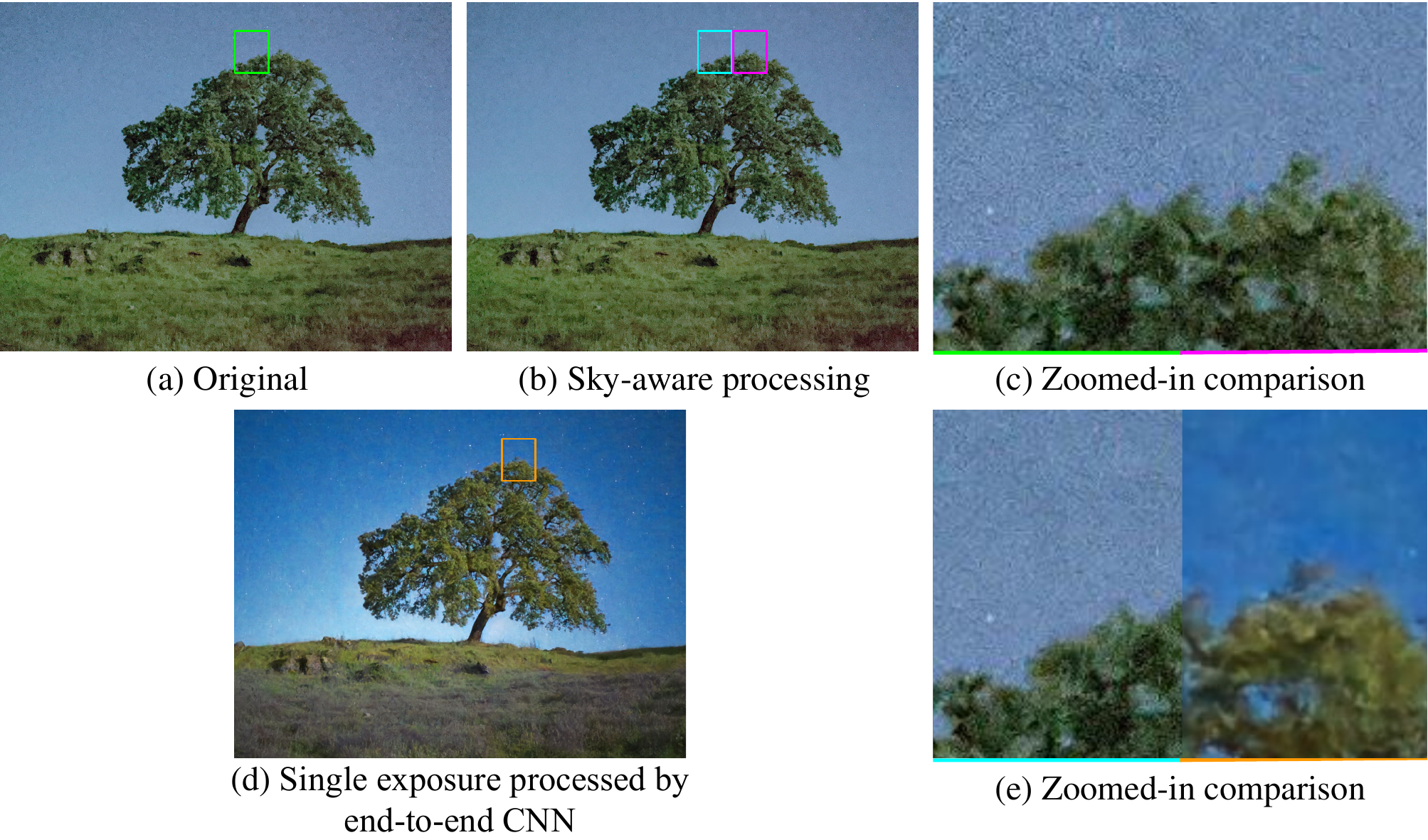}
\caption{Comparison between a low-light image, with and without sky-aware processing, and the end-to-end trained CNN described in \cite{chen2018learning}. A raw frame was captured with a Sony $\alpha$7S II, the same camera that was used to train the CNN, with an exposure time of 0.1 s. a) The original result. b) The result in (a) with sky-aware processing. (c) Zoomed-in regions of the original image (green) and sky-processed image (magenta). d) The result from the CNN. e) Zoomed-in regions of our result (cyan) and the result of the CNN (orange).}
\label{fig:noise_reduction_l2sitd}
\end{figure}

In this section we show an example of sky-aware noise reduction in a low-light image and compare it to an end-to-end convolutional neural network (CNN) \cite{chen2018learning} that produces a low-light image from a single raw frame. In \fig{fig:noise_reduction_l2sitd}a-c, our system is able to reduce the noise in the skies while preserving the details of the stars and the tree. 
\fig{fig:noise_reduction_l2sitd}d-e shows a comparison of our result to the result produced by the CNN of \cite{chen2018learning}. In this comparison, we used a raw frame captured with a similar camera model as was used for training the network in~\cite{chen2018learning} (Sony $\alpha$7S II). Because our white balancing algorithm was not calibrated for the Sony camera, and since color variation can affect the perception of details, we used Photoshop's automatic tool to color-match our results to the results produced by the CNN (as detailed in \cite{liba2019handheld}). Although the sky in the image produced by the CNN has less fine-grained noise compared to our result, our result has less low frequency noise in the skies and preserves more details of the foreground.

\section{Refining the annotations of the ADE20K dataset}
\label{sec:dataset_refinement}
In the Experimental Results Section, we chose the ADE20K dataset \cite{zhou2017scene} as our baseline dataset. We selected all the images that have the ``sky'' label and added to them $10\%$ random images without the skies. We do this for both the validations and training parts of the dataset so that in total we have 9187 images from the training set and 885 images from the validation set. We then refine this dataset with two different methods: 1) using the guided filter only (annotated as ADE20K+GF), and 2) using density estimation and the guided filter (annotated as ADE20K+DE+GF). 

Creating ADE20K+GF is straightforward: we input the annotated raw masks and the ADE20K images into the weighted guided filter algorithm (described in Section 3.1) with a confidence map of ones in each pixel. The parameters of the guided filter are $s=48$ and, $\lumasmooth =\chromasmooth=0.01$. In order to create ADE20K+DE+GF we had to create a heuristic for the ``undetermined'' label, in which we would apply inpainting using the density estimation algorithm described in Section 3.2.3. We used the following method: a) Run a Laplacian edge filter on the raw ADE20K sky masks to find the sky boundaries; b) Dilate the boundaries with an ellipse kernel with a radius of 4 pixels to generate the ``undetermined'' label; c) Add areas labeled as trees in ADE20K to our ``undertermined'' region. We do this because often the sky can be seen through tree branches and we have found that inpainting the entire tree to be more accurate than using raw annotations. Then, we inpaint the ``undetermined'' region using density estimation. For these experiments, we used a probability threshold of $p_c=0.97$. We then create the confidence map with values: $\cdetermined = 0.8$ (for the ``sky'' and ``not sky'' original raw labels), $\cinpainted = 0.6$  is for pixels inpainted as skies and $\cundetermined = 0.4$ for the remaining pixels. Finally, we apply the weighted guided filter, with parameters: $s=16$ and, $\lumasmooth =\chromasmooth=0.01$, and inputs: the inpainted annotations, the new confidence maps, and the original ADE20K images. Following the guided filter, in order to drive the intermediate mask values towards the edges of the range: $0$ and $1$, we applied sharpening to the mask, using \eq{eq:sigmoid_postprocessing}.

\begin{equation}
\label{eq:sigmoid_postprocessing}
S(x) = \frac{h(t_s(x-1/2)) - h(-t_s/2)}{h(t_s/2) - h(-t_s/2)}
\end{equation}

In which $x$ is the value of the sky mask, normalized to a range of $[0,1]$, $h(x) = 1/(1 + exp(-x))$ is the sigmoid function, and $t_s$ is the sharpness factor. We found that a sharpness factor, $t_s=15$ produces visually accurate masks.

\section{User study results}
\label{sec:user_study_results}
\begin{figure}
\includegraphics[width=\linewidth]{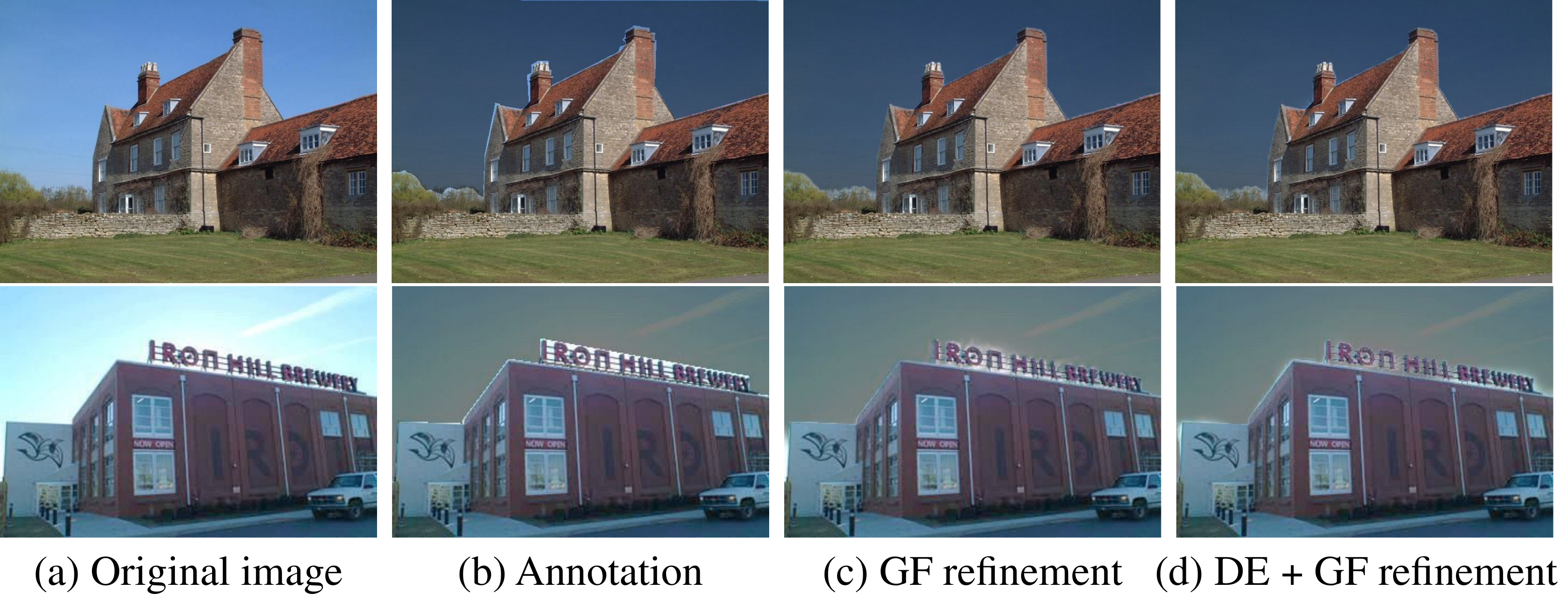}
\caption{Example images from the user study. In these images, the sky was darkened using either the original sky mask annotation of the ADE20K dataset or refined sky annotations. The user study evaluated the user's preference for images (b), (c), or (d). Sky darkening was applied to emphasize the shape of the sky mask, and is not an indication of our proposed darkening scheme which is described in Section 3.4 of the main paper}.
\label{fig:dataset_refinement_comparison}
\end{figure}

The goal of the user study is to show that raw binary annotations, demonstrated here using the ADE20K dataset \cite{zhou2017scene}, are too rough for computational photography applications. Because refined masks are more suitable for sky editing, a quantitative evaluation of mask accuracy should be performed using refined annotations as the ground-truth. The subjects were asked to rate which masks are more accurate: the raw annotations, the annotations refined only with the guided filter (GF) and the annotations refined with both density estimation and guided filter (DE+GF). Example images are shown in \fig{fig:dataset_refinement_comparison}. The annotations were evaluated one pair at a time. The images for the study were 100 randomly picked images with skies from the ADE20K validation set. We had 21 participants take the study. The results are in \tab{tab:user_study_results}. From the results we see that the users preferred masks refined with both DE and GF over only GF and any refinement was preferred over the raw annotations.

\begin{table}
  \centering
  \resizebox{\linewidth}{!}{
  \begin{tabular}{lccc}
                 & Raw & GF & DE+GF\\
  \toprule
  Raw versus GF       &  629 (30\%)& 1471 (70\%)&  \\
  Raw versus DE+GF & 365 (17.4\%) &  & 1735 (82.6\%)  \\
  GF versus DE+GF          & & 476 (22.7\%) & 1624 (77.3\%) \\
  Total       & 994 (15.8\%) & 1947 (30.9\%) & 3359 (53.3\%)  \\
  \bottomrule
  \end{tabular}
  }
  \captionsetup{justification=raggedright,singlelinecheck=false}
  \caption{The results of a user study comparing raw ground truth sky masks and refined sky masks from the ADE20K dataset.}
  \label{tab:user_study_results}
\end{table}

\section{Evaluation metrics}
\label{sec:evaluation_metrics}
Here we define the segmentation evaluation metrics used in our experiments. We use two metrics that take as input the binarized versions (at 0.5) of our ground-truth alpha mattes and of our predictions: mean intersection-over-union (mIOU) and misclassification rate (MCR):
\begin{align}
\operatorname{mIOU_{0.5}} &= \sum{\frac{TP}{TP+FP+FN}} \nonumber \\
\operatorname{MCR_{0.5}}  &= \sum{\frac{FP+FN}{M}}
\label{eq:binary_metrics}
\end{align}
where TP, FP and FN are the true positive, false positive, and false negative, respectively, and M is the number of pixels.
We also present a series of non-binarized error metrics: root mean square error (RMSE), mean absolute error (MAE), boundary loss (BL), and Jensen-Shannon Divergence (JSD): 
\begin{align}
\operatorname{RMSE}(X, Y) &= \sqrt{\frac{1}{M} \sum_i (X_i - Y_i)^2} \nonumber \\
\operatorname{MAE}(X, Y)  &= \frac{1}{M} \sum_i \left| X_i - Y_i \right| \nonumber \\
\operatorname{BL}(X, Y)  &= \sqrt{\frac{1}{M}\sum_i (\nabla X_i - \nabla Y_i )^2 \nonumber} \\
\operatorname{JSD}(X\parallel Y) &= \frac{1}{M} \sum_i \bigg( \frac{1}{2} \operatorname{KL} \left(X_i \parallel \frac{X_i+Y_i}{2} \right) + \nonumber \\
& \hphantom{= \frac{1}{M} \sum_i \bigg( }  \, \frac{1}{2} \operatorname{KL}\left(Y_i \parallel \frac{X_i+Y_i}{2} \right) \bigg)
\label{eq:continous_metrics}
\end{align}
Where $X$ and $Y$ are the predicted and true alpha mattes, $\nabla$ indicates the spatial gradient of an image, and $\operatorname{KL}(\cdot)$ is the KL divergence between two Bernoulli distributions (the true and predicted alpha matte at each pixel).

\begin{figure}
\includegraphics[width=\linewidth]{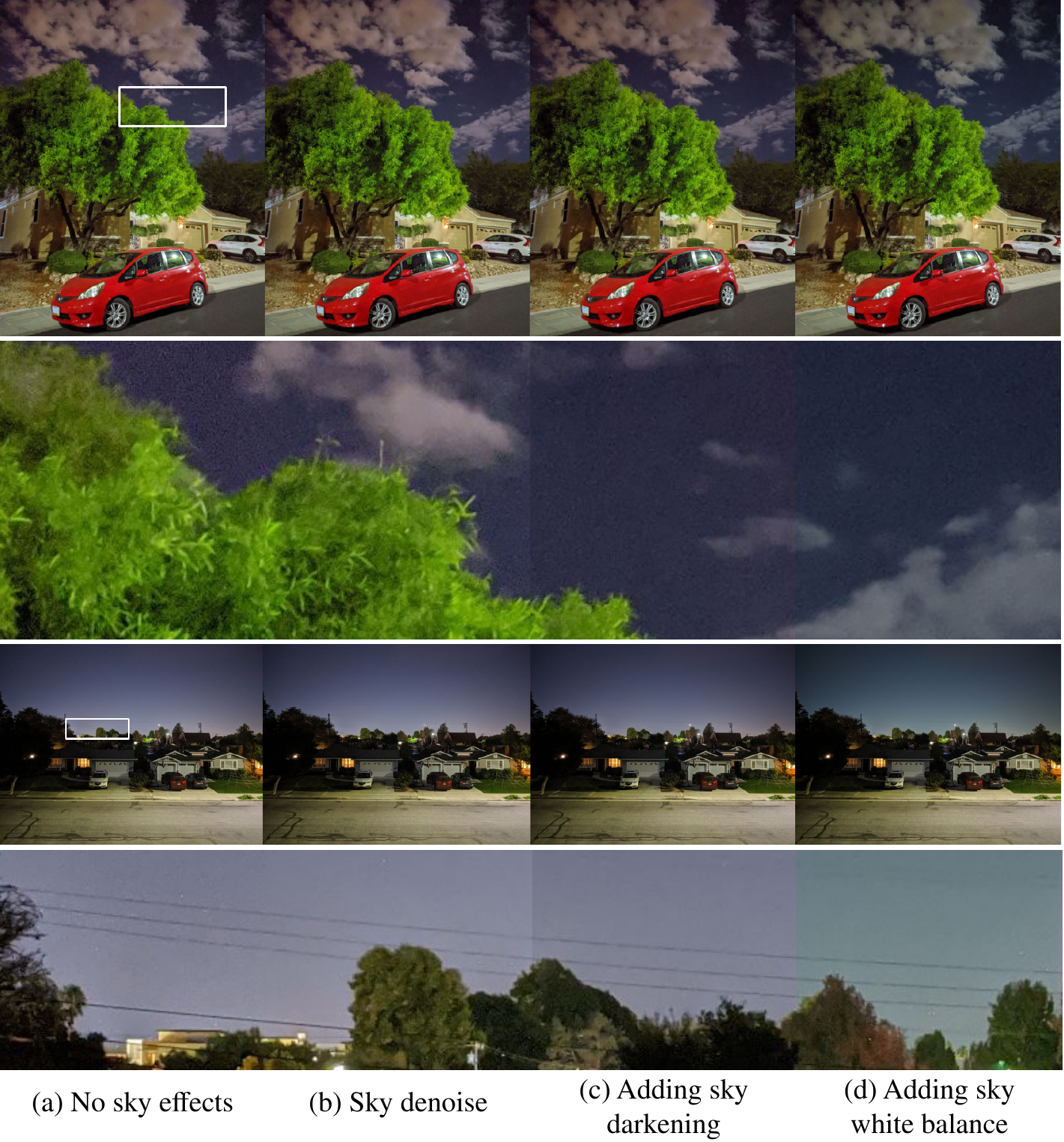}
\caption{The sky effects we propose, applied in sequence. Note that these effects are independent and do not rely on one another. a) The original image. b) Sky-specific noise reduction. c) Tonemapping applied to darken the sky. d) sky-inferred auto white balance gains applied according to the sky mask.}
\label{fig:sky_optimization_comparison}
\end{figure}

\begin{figure*}[t!]
    \centering
    \includegraphics[width=\textwidth]{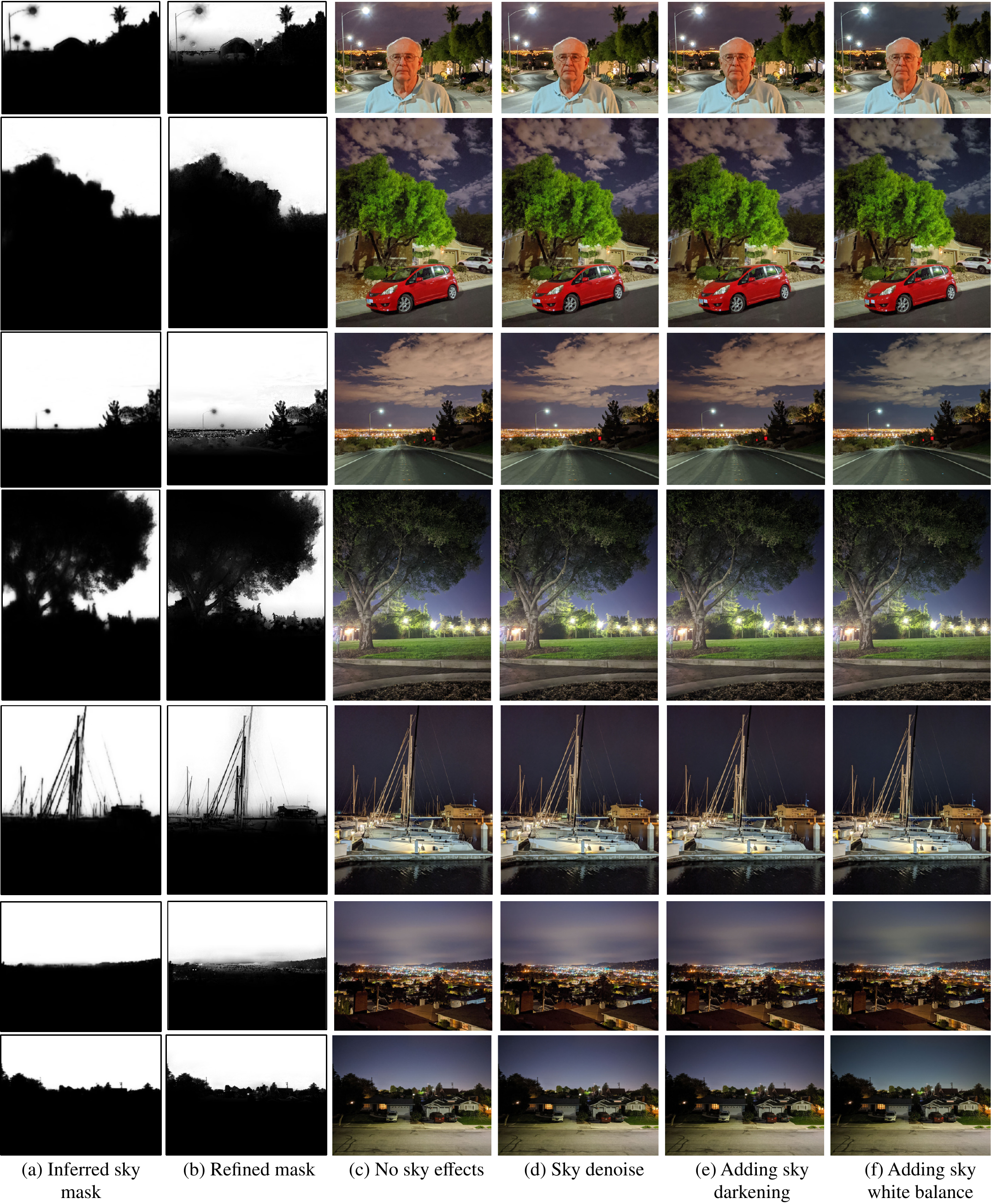}
    \caption{The sky affects applied one after the other. a) The inferred sky mask. b) The upsampled and refined mask, using our modified weighted guided filter. c) The original image, without sky effects. d) Sky-specific noise reduction is applied to the sky. The readers are encouraged to zoom-in to see the difference in noise characteristics. e) Tonemapping is applied to darken the sky. f) The sky-inferred auto white balance gains are applied to the sky pixels.}
    \label{fig:sup_step_by_step_results}
\end{figure*}

\section{Examples of the sky effects}
\label{sec:additional_examples}
The effects shown in \fig{fig:sky_optimization_comparison} and \fig{fig:sup_step_by_step_results} demonstrate the sky-aware processing steps performed by our pipeline. As shown in these figures, our procedure is able to accurately segment the sky and automatically improve its appearance for a variety of scenes. The effects are relatively subtle, as we have calibrated them to maintain the reliability of the scene and only alleviate the challenges of low-light imaging, without changing the image too much or potentially introducing new artifacts.